\documentclass[sigconf,nonacm]{acmart}

\setcopyright{none}
\acmDOI{}
\acmISBN{}
\acmConference{}{}{}
\acmBooktitle{}

\begin{document}

\title[Multi-Head ProbeRM]{Multi-Head ProbeRM: Rubric-Guided Reward Modeling from Frozen Model Representations}

\author{Chaokun Wang}
\affiliation{%
  \institution{University of Science and Technology of China}
  \city{Hefei}
  \country{China}
}
\email{wangck@mail.ustc.edu.cn}

\author{Zirui Liu}
\affiliation{%
  \institution{University of Science and Technology of China}
  \city{Hefei}
  \country{China}
}

\author{Zhenya Huang}
\affiliation{%
  \institution{University of Science and Technology of China}
  \city{Hefei}
  \country{China}
}

\author{Jinze Wu}
\affiliation{%
  \institution{iFLYTEK AI Research}
  \city{Hefei}
  \country{China}
}

\author{Suyi Liu}
\affiliation{%
  \institution{University of Science and Technology of China}
  \city{Hefei}
  \country{China}
}

\author{Junqiang Chen}
\affiliation{%
  \institution{University of Science and Technology of China}
  \city{Hefei}
  \country{China}
}

\author{Tingyue Pan}
\affiliation{%
  \institution{University of Science and Technology of China}
  \city{Hefei}
  \country{China}
}

\author{Shijin Wang}
\affiliation{%
  \institution{iFLYTEK AI Research}
  \city{Hefei}
  \country{China}
}

\begin{abstract}
  Reward models play a critical role in aligning large language models (LLMs) with human preferences. However, scalar reward models compress heterogeneous response qualities into one value, making the basis of a judgment difficult to inspect, while rubric-based generative judges make evaluation criteria explicit at the cost of high inference latency. We present Multi-Head ProbeRM, which learns five rubric-aligned dimension scores \emph{Accuracy, Compliance, Clarity, Coverage, and Fidelity} from offline teacher supervision and computes them from one frozen prompt--response encoding. The rubric defines the scoring targets during training; inference requires only the prompt and response, with no separate evaluation prompts, rubric generation, or teacher-model calls. We LoRA-adapt Qwen3 as a scalar reward model, freeze the backbone, train five bounded linear probes on teacher-annotated scores across 9,400 preference pairs, and rank responses by the arithmetic mean of the five outputs. Across RewardBench~2, JudgeBench, RM-Bench, and xf-edu-preference, Qwen3-4B achieves 68.19\%, 61.45\%, 72.96\%, and 61.50\%, for a four-benchmark mean of 66.02\%; Qwen3-8B achieves 69.51\%, 65.32\%, 73.00\%, and 64.40\%, for a mean of 68.06\%. On xf-edu-preference, both models exceed the strongest evaluated 4B--8B scalar and generative baselines at 56.50\% and 54.40\%. In a matched timing study, ProbeRM is 4.48$\times$ and 4.23$\times$ faster than OpenRubric on the 1,000-pair xf-edu-preference test set, and it scores the full four-benchmark suite in 1,351.23~s and 1,997.19~s at 4B and 8B, 17.52$\times$ and 20.65$\times$ faster than AutoRubric. These results show that a shared representation can deliver rubric-aligned dimension feedback and accurate preference ranking with efficient scoring.

\end{abstract}


\maketitle

\begin{figure}[!t]
    \centering
    \includegraphics[width=\linewidth]{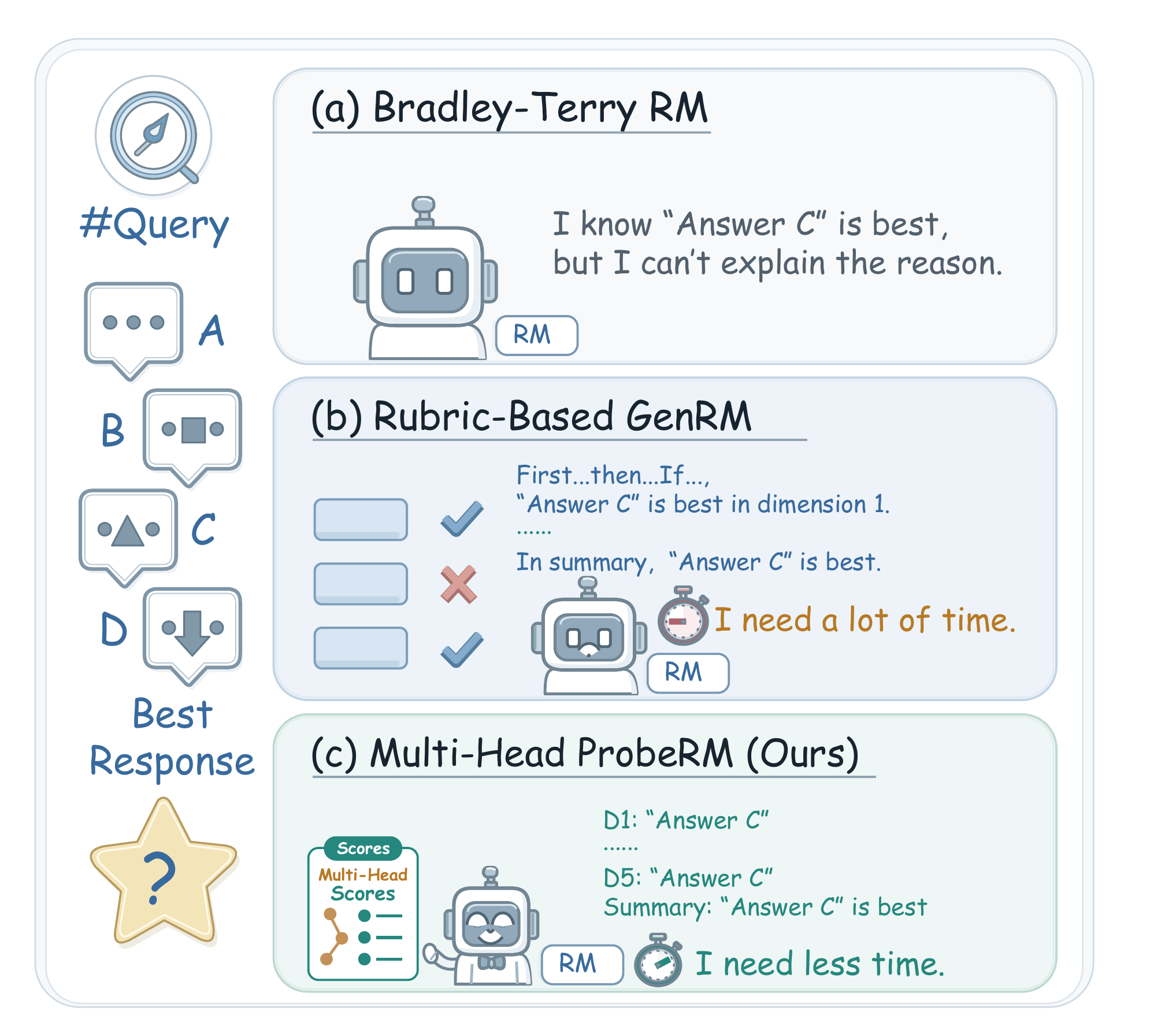}
    \caption{Comparison of reward-model outputs.} 
    \Description{Three approaches compare four candidate responses to a query: scalar reward scoring, rubric-based text generation, and direct prediction of five dimension scores with Multi-Head ProbeRM.}
    \label{fig:reward-model-comparison}
\end{figure}

\section{Introduction}
Reward modeling is central to aligning LLMs with human preferences through reinforcement learning~\cite{christiano2017deep,ouyang2022instructgpt}. As LLMs increasingly mediate access to information through conversational systems, response quality can rarely be reduced to a single notion of goodness—it spans distinct criteria such as factual accuracy, instruction compliance, and contextual fidelity. A practical evaluator must therefore rank candidates at scale while exposing the multi-dimensional basis of its judgments and keeping scoring latency manageable in practice.

Figure~\ref{fig:reward-model-comparison} contrasts three paradigms for response scoring. First, conventional scalar reward models compress response quality into a single score learned from preference comparisons. While sufficient for basic ranking, this scalar hides which quality dimensions determine a judgment: a fluent but inaccurate answer and a correct but incomplete one can receive similar scores for completely different reasons. This opacity prevents fine-grained error diagnosis and targeted model improvement, reflecting broader limitations of single-scalar proxies under optimization pressure~\cite{gao2023scaling,casper2023open}.

Rubric-based evaluation makes criteria explicit by decomposing response quality into natural-language requirements, yielding interpretable dimension-level judgments~\cite{gunjal2026rubrics,liu2026openrubrics}. Generative judges, however, must autoregressively generate and parse evaluation text for every candidate at test time, creating an inference bottleneck and inflating deployment cost as more dimensions are assessed. The central design question is how to retain rubric-level interpretability without token generation in the scoring loop.

Multi-attribute reward modeling and representation-based judging offer promising foundations for this goal. ArmoRM predicts broad human-preference attributes from frozen representations and learns a prompt-conditioned gating network for reward aggregation~\cite{wang2024armorm}. INSPECTOR extracts aspect-level evaluation signals via aspect-specific prompts and multi-layer hidden states~\cite{li2026representation}, while Xie et al.\ train probes over frozen representations to verify instance-specific rubric criteria~\cite{xie2026small}. However, existing representation-based approaches either rely on coarse, non-rubric attributes or still require dimension- and instance-specific prompts at test time, preventing plug-and-play reward computation from a single standard forward pass.

We propose Multi-Head ProbeRM, which distills a shared evaluation rubric—comprising 25 reusable criteria organized into Accuracy, Compliance, Clarity, Coverage, and Fidelity—into five parallel scoring heads over a single prompt--response representation. Because all five heads read the same hidden representation directly, ProbeRM produces both dimension-level diagnostics and an aggregated scalar reward in a single forward pass, eliminating test-time rubric generation or aspect-specific prompting.

To endow the shared representation with strong discriminative power before probe fitting, we adopt a two-stage training pipeline. First, we LoRA-adapt Qwen3~\cite{yang2025qwen3,hu2021lora} as a standard scalar reward model using a Bradley--Terry objective on large-scale preference data, then freeze the adapted backbone. Second, we annotate 10,000 preference pairs with dual-teacher dimension scores and fit five bounded linear probes (on 9,400 training pairs) over multi-layer token representations. At inference, the mean of the five probe outputs serves directly as the final ranking score.

We evaluate Multi-Head ProbeRM at two Qwen3 scales (4B and 8B) on three public benchmarks—RewardBench~2, JudgeBench, and RM-Bench~\cite{malik2025rewardbench2,tan2024judgebench,liu2025rmbench}—and the human-preference xf-edu-preference set, measuring pairwise ranking accuracy and end-to-end scoring latency under a matched timing protocol. Across all four evaluations, a single frozen representation yields rubric-level diagnostics, competitive ranking accuracy, and markedly lower inference cost than generative rubric judges.

Our main contributions are summarized as follows:
\begin{itemize}
  \item \textbf{Rubric quality without rubric inference.} Five linear probes over one frozen prompt--response encoding distill a 25-criterion rubric into dimension-level scores and a scalar ranking reward in a single forward pass—no rubric generation, aspect prompting, or teacher calls at test time.
  \item \textbf{Dimension scores that are causally faithful.} Each probe drops sharpest under its own targeted degradation and correlates most strongly with its matching teacher dimension, showing that the shared representation disentangles the five criteria.
  \item \textbf{Better rankings at a fraction of the cost.} ProbeRM averages 66.02\% and 68.06\% across four benchmarks at 4B and 8B, surpasses the strongest 4B--8B baselines on xf-edu-preference by a clear margin, and scores the full suite 17.52$\times$ and 20.65$\times$ faster than AutoRubric~\cite{xie2026autorubric}.
\end{itemize}

\begin{figure*}[!tp]
  \centering
  \includegraphics[width=\textwidth]{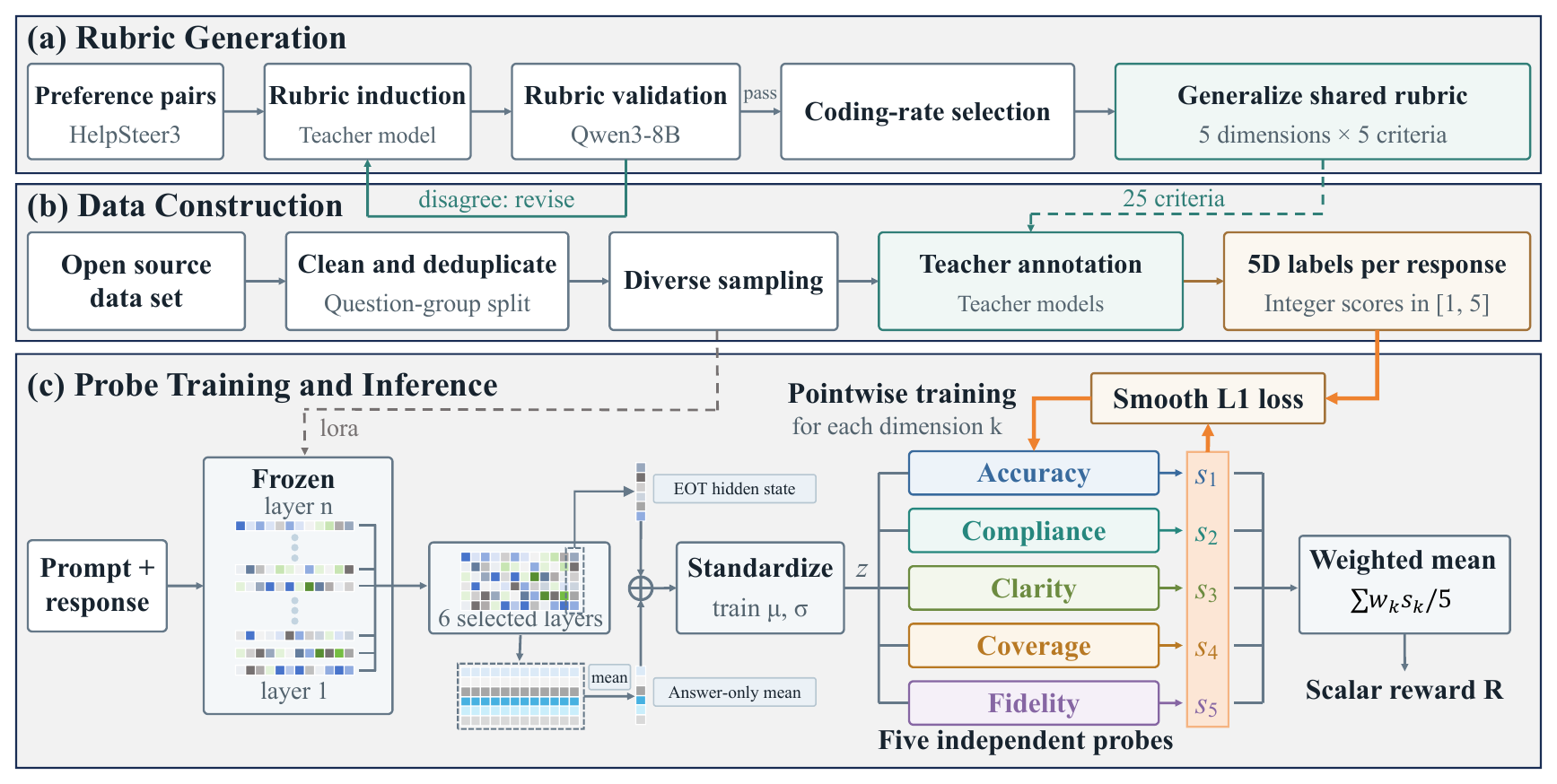}
  \caption{Overview of Multi-Head ProbeRM. (a) GPT-6.1-Sol generates and revises candidate rubrics, and Qwen3-8B validates them. After coding-rate compression, GPT-6.1-Sol generalizes the retained rubrics into 25 criteria across five dimensions. (b) DeepSeek-4.1-flash and GPT-6.1-Sol each assign integer scores in $[1,5]$; their per-dimension mean forms the training targets, which can take half-integer values. (c) Five independent probes learn from these targets using frozen LoRA-adapted Qwen3 features. The aggregation weights are all $w_k=1$, so the scalar reward is the arithmetic mean of the five outputs.} \Description{Three stages show offline rubric generation, dual-teacher annotation, and probe training and inference. A shared rubric contains five dimensions with five criteria each. Two teachers provide integer scores, which are averaged without rounding to form training targets. A frozen LoRA-adapted Qwen3 encoder produces six-layer end-of-turn and answer-only mean features. The concatenated standardized feature feeds independent Accuracy, Compliance, Clarity, Coverage, and Fidelity probes, trained with Smooth L1 loss. Their continuous outputs are averaged with all aggregation weights set to one to obtain a scalar reward.}
  \label{fig:five-dimension-probe}
\end{figure*}

\section{Related Works}
\subsection{Multi-Attribute Reward Modeling}
Conventional scalar reward models compress human preferences into a single value~\cite{ouyang2022instructgpt}, whereas direct alignment methods bypass explicit reward modeling altogether via an implicit policy parameterization~\cite{rafailov2023direct}, leaving no standalone, inspectable scorer. To make quality dimensions explicit during training, HelpSteer and HelpSteer2 introduce pointwise annotations across multiple response attributes~\cite{wang-etal-2024-helpsteer,wang2024helpsteer2}. Building on frozen representations from a Bradley--Terry backbone, ArmoRM fits linear readouts for 19 attributes pooled from disparate datasets and learns a prompt-conditioned mixture-of-experts (MoE) gate to aggregate them for ranking~\cite{wang2024armorm}. While sharing the principle of probing frozen reward representations, Multi-Head ProbeRM departs from ArmoRM in both supervision and aggregation: rather than combining heterogeneous dataset attributes through a learned gating network, ProbeRM distills a unified, preference-derived rubric into five standardized dimensions via dual-teacher supervision and aggregates them with a simple arithmetic mean. This design evaluates every response under a consistent set of criteria and renders each dimension score $s_k$ directly attributable as an exact $s_k/5$ contribution to the final scalar reward.

\subsection{Rubric-Guided Reward Modeling}
LLM-as-a-Judge uses prompted language models to evaluate open-ended responses~\cite{zheng2023judging}. To make evaluation criteria explicit, subsequent methods condition judges on structured rubrics: Prometheus and Prometheus~2 generate natural-language feedback alongside scores from user-specified rubrics~\cite{kim2024prometheus,kim2024prometheus2}, while Rubrics as Rewards constructs task-specific criteria for open-ended reinforcement learning~\cite{gunjal2026rubrics}. Moving beyond manual rubric crafting, OpenRubrics automates rubric induction from preference data to train rubric-conditioned evaluators~\cite{liu2026openrubrics}, and Auto-Rubric systematically induces, validates, revises, and compresses criteria for prompted judging~\cite{xie2026autorubric}. Despite these advances, existing rubric-guided judges keep the rubric inside the inference loop, requiring rubric conditioning and autoregressive text generation at test time. In this work, we adopt Auto-Rubric's induction and compression procedure purely offline to construct a unified rubric, and train Multi-Head ProbeRM to distill the resulting dimension scores into representation readouts. This shifts rubric usage from runtime prompt conditioning into offline supervision, yielding both dimension-level feedback and a scalar ranking reward directly from the prompt and response alone.

\subsection{Representation-Based Reward Modeling}
Representation-based judges read evaluation signals directly from hidden states rather than autoregressively decoding a textual verdict. INSPECTOR predicts teacher-provided aspect scores for reference-free evaluation and data filtering by constructing aspect-specific prompts and pooling features across selected layers~\cite{li2026representation}. Similarly, Xie et al.\ train frozen probes over representations conditioned on a prompt, response, and individual rubric criterion, aggregating criterion-satisfaction probabilities into rewards for group relative policy optimization (GRPO)~\cite{xie2026small,shao2024deepseekmath}. While these methods avoid text generation, conditioning the backbone on each aspect prompt or criterion still requires a separate forward pass per evaluation target. Multi-Head ProbeRM eliminates this per-criterion overhead by mapping a single shared representation of the prompt--response pair $(q,y)$ to five continuous rubric scores in parallel. All heads reuse the exact same backbone encoding, amortizing fine-grained rubric evaluation and scalar ranking into a single forward pass.

\section{Our Method}

Figure~\ref{fig:five-dimension-probe} illustrates the workflow of Multi-Head ProbeRM across three stages: constructing a shared rubric of 25 criteria organized into five dimensions, generating offline supervision by averaging integer scores from two teacher models, and fitting five linear probes whose mean yields the final reward. While the rubric defines the semantic target of each head during training, the adapted and frozen backbone encodes only the prompt and response. This design cleanly decouples the offline specification of evaluation criteria from their joint, single-pass computation at inference.

\subsection{Rubric Generation}
\label{sec:rubric-generation}
\begingroup
\setlength{\emergencystretch}{2em}

To ensure that each evaluation dimension maintains a consistent semantic meaning across diverse queries, all probe heads share a unified definition of response quality. We construct this standardized rubric offline by applying the preference-conditioned induction and compression procedure of Auto-Rubric~\cite{xie2026autorubric} to our preference training data, yielding a reusable hierarchy of dimensions and fine-grained criteria.

\paragraph{Preference-conditioned induction and validation.}
From the General domain of the HelpSteer3-Preference training split~\cite{wang2025helpsteer3preference}, we filter out ties, empty responses, and duplicate pairs, and sample 70 preference pairs with distinct queries. Each instance $(q_i, y_i^{+}, y_i^{-})$ consists of a query, a preferred response, and a rejected response. Given each pair, GPT-6.1-Sol~\cite{openai2026gpt61sol} proposes a candidate rubric, which is then validated by a separate Qwen3-8B judge~\cite{yang2025qwen3} tasked with ranking the two responses in a seeded randomized order (to mitigate position bias~\cite{zheng2023judging}) without access to the ground-truth preference label. If the judge produces a valid strict ranking that aligns with the preference label, the rubric is retained; otherwise, GPT-6.1-Sol iteratively revises the rubric based on the judge's feedback for re-evaluation. Allowing up to ten attempts per pair, this loop yields 68 validated rubrics, with two pairs reaching the attempt limit.

\paragraph{Coding-rate compression.}
After Unicode and whitespace normalization, we remove exact duplicate rubrics and embed each unique rubric into a $d=1{,}024$-dimensional space using \texttt{qwen3.7-text-embedding-}\allowbreak\texttt{flash}~\cite{qwen2026textembedding}. For any nonempty subset $\mathcal{S}$ of size $n = |\mathcal{S}|$, let $E_{\mathcal{S}}\in\mathbb{R}^{d\times n}$ denote the matrix whose columns are the embeddings of the rubrics in $\mathcal{S}$. We quantify the semantic diversity preserved by $\mathcal{S}$ via the lossy subspace coding rate~\cite{ma2007segmentation}:
\begin{equation}
  \mathcal{C}(\mathcal{S}) = \frac{1}{2}\log\det\!\left(
    I_d + \frac{E_{\mathcal{S}}E_{\mathcal{S}}^{\top}}{\epsilon^2 n}
  \right),
  \label{eq:rubric-coding-rate}
\end{equation}
where $\epsilon=0.5$ and $\mathcal{C}(\emptyset)=0$. Following Auto-Rubric~\cite{xie2026autorubric}, we process the deduplicated rubric pool in batches of ten, greedily adding the rubric that maximizes the marginal coding-rate gain $\mathcal{C}(\mathcal{S}\cup\{r\}) - \mathcal{C}(\mathcal{S})$ until the gain falls below $0.002$ or the batch is exhausted. This compression yields a compact core of 29 rubrics.

\paragraph{Generalization into shared dimensions.}
GPT-6.1-Sol generalizes the 29 core rubrics into 25 reusable criteria, organized evenly (five per dimension) across the following five evaluation dimensions:
\begin{itemize}
  \item \emph{Factual Accuracy \& Context Consistency (D1)}: verifies factual correctness and consistency with supplied evidence.
  \item \emph{Strict Compliance with Requirements (D2)}: checks adherence to explicit instructions, constraints, and formatting rules.
  \item \emph{Clarity \& Structured Presentation (D3)}: assesses readability, logical organization, and conciseness.
  \item \emph{Comprehensive \& Coherent Coverage (D4)}: evaluates completeness and depth in addressing all parts of the query.
  \item \emph{Narrative \& Context Fidelity (D5)}: measures stylistic continuity, tone alignment, and faithful preservation of contextual nuance.
\end{itemize}
The complete 25-criterion rubric is provided in Appendix~\ref{app:general-rubrics} and establishes a unified semantic definition across all queries. In the subsequent offline annotation stage, both teacher models—GPT-6.1-Sol and DeepSeek-4.1-flash~\cite{deepseek2026api}—score responses against these same five dimensions (`D1--D5`).

\endgroup

\subsection{Data Construction}
\label{sec:data-construction}

Each supervision instance is formatted as a preference triple $(q, y^{+}, y^{-})$ comprising the complete conversational context $q$, a preferred response $y^{+}$, and a rejected response $y^{-}$, with all original text and multi-turn message ordering preserved verbatim. We construct two datasets from these preference pairs: a large-scale preference corpus for Stage~1 backbone adaptation and a dimension-annotated subset for Stage~2 probe training.

\paragraph{Source cleaning and split isolation.}
\begingroup
\setlength{\emergencystretch}{2em}
We pool Skywork-Reward-Preference-80K-v0.2 and HelpSteer3-Preference~\cite{liu2024skywork,wang2025helpsteer3preference}, containing 77,016 and 40,476 raw preference pairs, respectively. We filter out ties, empty or invalid responses, normalized-identical response pairs, exact duplicates, and duplicate instances with conflicting preference labels, while also quarantining any overlaps with evaluation benchmarks and historical hold-out sets. Deduplication and overlap matching apply NFKC normalization, Unicode case folding, and whitespace collapsing solely for identity comparison, leaving the raw model inputs unmodified. The resulting cleaned pool comprises 96,855 pairs. To prevent data leakage across training stages, we group instances by the normalized final non-empty user turn and partition them at the question-group level into 95,017 training pairs (80,869 groups) and 1,838 development pairs (1,641 groups). Stage~1 LoRA adaptation uses all 95,017 training pairs; Stage~2 probe training pairs are sampled strictly from the training split, whereas the 600-pair probe validation set is drawn from the development split, ensuring zero question-group overlap with either training stage.
\endgroup

\paragraph{Probe-training sample.}
To construct the Stage~2 dataset, we curate 10,000 unique preference pairs comprising 9,400 training pairs drawn from the 95,017-pair training split and the fixed 600-pair validation set from the development split. Stratified sampling balances data source, topic, response length, and language without access to teacher annotations, while filtering out degenerate marker-only contexts and ensuring prompt uniqueness. The 600 validation pairs are held out solely for monitoring probe validation loss and checkpoint selection.

\paragraph{Dual-teacher annotation and target formation.}
Both the probe-training and validation splits are annotated under the shared five-dimension rubric by two teacher models: DeepSeek-4.1-flash~\cite{deepseek2026api} and GPT-6.1-Sol. Each teacher independently assigns an integer score in $\{1,2,3,4,5\}$ to every response across all five dimensions. We align the two teacher exports by exact matching on the context, chosen response, and rejected response. For response $i$ and dimension $k$, we compute the ground-truth target as their unrounded arithmetic mean:
\begin{equation}
  t_{ik}=\frac{d_{ik}+g_{ik}}{2},
  \label{eq:mean-teacher-target}
\end{equation}
where $d_{ik}$ and $g_{ik}$ denote the integer scores from DeepSeek-4.1-flash and GPT-6.1-Sol, respectively. Retaining this unrounded mean places the training targets on a fine-grained $0.5$-point grid in $[1,5]$ (as illustrated in Figure~\ref{fig:five-dimension-probe}(b)--(c), where individual teacher integer scores are averaged to supervise the probes). Across the 10,000 preference pairs, this process yields 20,000 scored responses and 100,000 dimension-level supervision targets.

\subsection{Reward Model Training and Inference}
\label{sec:reward-model}

\paragraph{LoRA adaptation with pairwise outcome rewards.}
In Stage~1, we adapt Qwen3-4B and Qwen3-8B into scalar outcome reward models via LoRA on the 95,017 training pairs from Section~\ref{sec:data-construction}. For each prompt--response pair $(q,y)$, the adapted model outputs a scalar logit $r_{\theta}(q,y)$, where $\theta$ denotes the trainable LoRA and scalar-head parameters. Given a preference triple $(q, y^{+}, y^{-})$ with $y^{+}\succ y^{-}$, we minimize the Bradley--Terry pairwise loss~\cite{bradley1952rank}:
\begin{equation}
  \begin{aligned}
    &\mathcal{L}_{\mathrm{BT}}(q,y^{+},y^{-}) \\
    &\quad = -\log\operatorname{sigmoid}\!\left(
      r_{\theta}(q,y^{+})-r_{\theta}(q,y^{-})\right),
  \end{aligned}
  \label{eq:lora-bt-loss}
\end{equation}
averaged over the training set. Once adaptation completes, we freeze both the base and LoRA weights, preserving the preference-aligned backbone as a fixed feature extractor for Stage~2.

\paragraph{Frozen representation extraction and probe parameterization.}
Using the frozen adapted backbone, we extract a multi-layer representation for each prompt--response pair $(q,y)$ with thinking mode disabled and a maximum sequence length of 4,096 tokens (truncating overlength inputs to keep the first 512 and last 3,584 tokens). Across six selected layers $\mathcal{L}=\{10,15,21,26,31,36\}$, we concatenate the final assistant end-of-turn state $h_{\ell}^{\mathrm{eot}}$ (falling back to the last valid token if truncated) and the mean state over answer-token positions $\mathcal{A}$ (excluding prompt, end-of-turn, and padding tokens):
\begin{equation}
  \begin{aligned}
    x(q,y) &= \big[ h_{\ell}^{\mathrm{eot}} \, ; \,
    \bar{h}_{\ell}^{\mathrm{ans}} \big]_{\ell \in \mathcal{L}}
    \in \mathbb{R}^{D}, \\
    \bar{h}_{\ell}^{\mathrm{ans}} &=
    \frac{1}{|\mathcal{A}|}\sum_{t \in \mathcal{A}} h_{\ell,t},
  \end{aligned}
  \label{eq:probe-feature}
\end{equation}
where $H$ denotes the hidden size and $D=12H$, yielding $D=30{,}720$ for Qwen3-4B and $D=49{,}152$ for Qwen3-8B.

Next, we standardize each feature coordinate $j\in\{1,\ldots,D\}$ using the population mean $\mu_j$ and standard deviation $\sigma_j$ computed strictly over the 18,800 training responses (from the 9,400 probe-training pairs):
\begin{equation}
  z_j(q,y)=\frac{x_j(q,y)-\mu_j}{\max(\sigma_j,10^{-4})}.
  \label{eq:probe-normalization}
\end{equation}
The resulting statistics $(\mu, \sigma)$ are fixed and applied unchanged to all validation and benchmark inputs.

Finally, for each dimension $k\in\{1,\ldots,5\}$ (ordered as D1--D5), we parameterize an independent bounded linear probe~\cite{alain2016understanding,belinkov2022probing}:
\begin{equation}
  \begin{aligned}
    s_k(q,y) &= 1+4\,\operatorname{sigmoid}
    \!\left(\mathbf{w}_k^{\top}z(q,y)+b_k\right) \in [1,5],
  \end{aligned}
  \label{eq:probe-score}
\end{equation}
where $\mathbf{w}_k\in\mathbb{R}^D$ and $b_k\in\mathbb{R}$ are the trainable weight vector and bias of head $k$. All five heads share the standardized feature vector $z(q,y)$ but are trained and checkpoint-selected independently per dimension.

\paragraph{Pointwise training with mean targets.}
For each head $k$, we train on the set of 18,800 individual training responses $\mathcal{D}_{\mathrm{tr}}$ by minimizing the Smooth L1 loss~\cite{huber1964robust} against the dual-teacher target $t_{ik}$:
\begin{equation}
  \mathcal{L}_k =
  \frac{1}{|\mathcal{D}_{\mathrm{tr}}|}
  \sum_{i\in\mathcal{D}_{\mathrm{tr}}}
  \ell_{\mathrm{SL1}}\!\left(s_k(q_i,y_i),t_{ik}\right).
  \label{eq:probe-loss}
\end{equation}
Each head is optimized using full-batch AdamW~\cite{loshchilov2019decoupled} across a six-configuration grid combining learning rates $\{10^{-4},3\times10^{-4},10^{-3}\}$ and weight decays $\{10^{-4},10^{-2}\}$. We initialize the weights to zero and set the bias to the logit-transformed training-set mean target. Training runs for up to 600 epochs with early stopping (patience 60, minimum validation-loss improvement $10^{-7}$), retaining for each head the checkpoint that achieves the lowest Smooth L1 loss on the validation set.

\paragraph{Inference and scalar reward.}
At test time, evaluating a candidate pair $(q,y)$ requires only a single forward pass through the frozen backbone, followed by multi-layer pooling, coordinate standardization, and parallel evaluation of the five linear probes. Setting the dimension aggregation weights in Figure~\ref{fig:five-dimension-probe}(c) to equal unity ($\alpha_k=1$, or $w_k=1$ if matching the figure notation), we compute the final scalar reward as
\begin{equation}
  R(q,y)=\frac{1}{5}\sum_{k=1}^{5}s_k(q,y).
  \label{eq:scalar-reward}
\end{equation}
This single-pass procedure simultaneously exposes the five dimension scores $\{s_k(q,y)\}_{k=1}^{5}$ for fine-grained diagnostic inspection and provides $R(q,y)$ for direct pairwise preference ranking.

\section{Experiments}

\subsection{Datasets and Experiment Settings}
\label{sec:experiment-settings}

\paragraph{Training data.}
Following the cleaning and question-grouped splitting protocol in Section~\ref{sec:data-construction}, Stage~1 LoRA adaptation trains on all 95,017 preference pairs (with 1,838 pairs held out as the source development set). Stage~2 probe training uses the 10,000 dual-teacher annotated pairs, split into 9,400 training pairs (18,800 responses) and 600 validation pairs (1,200 responses).

\paragraph{Model and optimization settings.}
We evaluate Qwen3-4B and Qwen3-8B after LoRA adaptation as scalar outcome reward models on the full 95,017-pair training split using Equation~\ref{eq:lora-bt-loss}. The adapted backbones are frozen, and the five linear probes are trained on cached hidden-state features using the Smooth L1 objective and optimization settings of Section~\ref{sec:reward-model}. Each head is selected independently by its lowest development loss; benchmark scores are never used for selection. The maximum input length is 4,096 tokens, with a 512-token prefix and the remaining tail retained for longer inputs, and the fixed mean of the five predicted scores is used for ranking.

\paragraph{Hardware.}
The experimental server is equipped with two NVIDIA V100-SXM2 GPUs, each with 32\,GB of device memory, 24 vCPUs on an Intel Xeon Platinum 8255C processor at 2.50\,GHz, and 86\,GB of system memory.

\subsection{Performance of Multi-Head ProbeRM}
\subsubsection{Benchmarks}
\label{sec:benchmarks}
We evaluate response ranking across four benchmarks spanning diverse evaluation aspects: general candidate selection (RewardBench~2), factual and reasoning correctness (JudgeBench), style-controlled robustness (RM-Bench), and domain-specific educational preferences (\texttt{xf-edu-preference}). Multi-Head ProbeRM scores each candidate response via the scalar reward $R(q,y)$ in Equation~\ref{eq:scalar-reward}. Following standard protocol, we report the official aggregate accuracy for each benchmark, along with their unweighted mean (\textsc{Avg}) across all four benchmarks.

\emph{RewardBench 2}~\cite{malik2025rewardbench2} contains 1,865 evaluation groups across Factuality, Precise Instruction Following, Math, Safety, Focus, and Ties. The first five categories require the correct response to outrank three incorrect candidates. The Ties category measures how scores distinguish incorrect responses while treating equally valid responses consistently.

\emph{JudgeBench}~\cite{tan2024judgebench} tests whether a judge can distinguish a correct response from an incorrect one on knowledge, reasoning, mathematics, and coding problems. We report the overall score over 620 pairs, combining the GPT and Claude subsets with the benchmark's official single-game aggregation.

\emph{RM-Bench}~\cite{liu2025rmbench} evaluates sensitivity to subtle content differences under changes in response length and formatting. It contains 1,327 prompts across Chat, Code, Math, and Safety. Each prompt has three preferred and three rejected responses in concise, detailed plain-text, and detailed Markdown styles, yielding 11,943 pairwise comparisons. We report the overall score aggregated across the Easy, Normal, and Hard conditions and the four domains using the benchmark's averaging scheme.

\emph{xf-edu-preference} contains 1,000 human-preference pairs comprising 2,000 candidate responses from real educational applications. The four scenarios are Chinese language arts AI analysis, Chinese language arts interactive video, English AI analysis, and English interactive video, with 250 pairs per scenario. Both the questions and candidate responses originate from actual application use, and all evaluation inputs are dialogue text, including those from the interactive-video scenarios. Professional teachers with relevant subject expertise select the preferred response under established annotation rules, with model identities hidden and candidate order randomized to suppress the position bias reported for LLM judges~\cite{zheng2023judging}. Each pair has one preferred response. We report the number of correct selections divided by all 1,000 pairs; predicted ties, refusals, and unparseable judgments count as errors. These pairs are reserved for final testing and are also used for the timing evaluation. The labels assess overall preference ranking rather than individual rubric dimensions. Appendix~\ref{app:edu-preference} details the annotation, scoring, and test-isolation protocol.

All benchmark scores are reported as percentages. The AVG column in Table~\ref{tab:baseline-performance} is the arithmetic mean of RewardBench 2, JudgeBench, RM-Bench, and xf-edu-preference.

\subsubsection{Baselines}
\label{sec:baselines}
We compare Multi-Head ProbeRM with general-purpose and rubric-based generative judges, learned reasoning judges, and pretrained scalar reward models in Table~\ref{tab:baseline-performance}. Controls for LoRA adaptation and probe design use the same Qwen3 backbones and data splits and are
reported separately in Table~\ref{tab:ablation}.

\emph{General and rubric-based judges.}
The general-purpose judges are GPT-4o~\cite{openai2024gpt4osystemcard}, Claude-3.5-Sonnet~\cite{anthropic2024claude35sonnet}, Qwen3-32B~\cite{yang2025qwen3}, Qwen3.8-flash-next~\cite{qwen2026design,qwen3.8flashnext}, and GLM5.3-flash~\cite{glm2026flash}. We also evaluate Qwen3-4B and Qwen3-8B using the official prompt for each benchmark, labeled ``official prompts'' in the result tables. We evaluate AutoRubric~\cite{xie2026autorubric} with Qwen3-4B and Qwen3-8B, using the fixed HelpSteer3 rubric prompt to rank candidate responses jointly. OpenRubric-4B and OpenRubric-8B~\cite{liu2026openrubrics} first generate a rubric from the request, then use a rubric-conditioned judge to compare responses. The rubric-based methods apply explicit natural-language criteria during inference.

\emph{Learned reasoning judges.}
We include RM-R1-DeepSeek-Distilled-Qwen-7B and RM-R1-Qwen-Instruct-7B~\cite{chen2025rmr1}, which learn rubric-guided evaluation through reasoning-chain distillation and reinforcement learning, and JudgeLRM-7B~\cite{chen2025judgelrm}, which trains a reasoning judge with outcome-driven rewards. These baselines compare our offline rubric supervision with learned generative evaluation.

\emph{Pretrained scalar reward models.}
We compare with Llama-3-OffsetBias-RM-8B~\cite{park2024offsetbias}, InternLM2-7B-Reward~\cite{cai2024internlm2}, Skywork-Reward-Llama-3.1-8B-v0.2~\cite{liu2024skywork} (abbreviated as Skywork-8B-v0.2), and ArmoRM-8B-v0.1~\cite{wang2024armorm}. These models assign a scalar reward to each prompt--response pair for ranking. ArmoRM combines multiple reward objectives through a gating network, providing a comparison with our fixed averaging of dimension scores.

\subsubsection{Performance}
\label{sec:performance}
Table~\ref{tab:baseline-performance} summarizes performance across the four benchmarks. Qwen3-4B and Qwen3-8B reach AVG scores of 66.02\% and 68.06\%, respectively. Scaling from 4B to 8B improves RewardBench 2 by 1.32 points, JudgeBench by 3.87 points, RM-Bench by 0.04 points, xf-edu-preference by 2.90 points, and AVG by 2.04 points. On xf-edu-preference, the two models score 61.50\% and 64.40\%, above the strongest scalar baseline in the table (Skywork-8B-v0.2 at 56.50\%) and the strongest listed 4B--8B generative baseline (Qwen3-8B with official prompts at 54.40\%). Both models also exceed the 60.10\% score of GLM5.3-flash, the best-performing general-purpose judge on this test set.

\paragraph{Comparison with rubric-based judges.}
Against OpenRubric, Multi-Head ProbeRM improves JudgeBench, RM-Bench, and xf-edu-preference by 19.68, 4.73, and 11.10 points at 4B, and by 25.16, 4.73, and 11.90 points at 8B, respectively. Against AutoRubric, the corresponding gains are 43.39, 18.73, and 10.60 points at 4B, and 12.58, 10.86, and 13.90 points at 8B. OpenRubric results are unavailable for RewardBench 2, so we restrict these comparisons to the reported benchmarks. The gains on RM-Bench show that the learned dimension scores remain useful under changes in response style while avoiding autoregressive judging.

\paragraph{Effect of backbone scale.}
The 4B-to-8B improvements are consistent across all four benchmarks, with the largest gain on JudgeBench, followed by xf-edu-preference. This scaling trend indicates that the probe readout can exploit the additional evaluative information in the larger LoRA-adapted representation.

\begin{table*}[!t]
\centering
\small
\setlength{\tabcolsep}{4pt}
\renewcommand{\arraystretch}{1.12}

\resizebox{\textwidth}{!}{%
\begin{tabular}{lccccc}
\toprule
Model
& \shortstack[c]{RewardBench 2(\%)}
& \shortstack[c]{JudgeBench(\%)}
& \shortstack[c]{RM-Bench(\%)}
& \shortstack[c]{xf-edu-preference(\%)}
& \shortstack[c]{AVG(\%)} \\
\midrule

\multicolumn{6}{l}{\textbf{General-purpose LLM judges}} \\[-2pt]

GPT-4o
& 64.90 & 59.80 & 73.10 & -- & -- \\

Claude-3.5-Sonnet
& 64.70 & 64.80 & 74.50 & -- & -- \\

Qwen3-32B
& 64.89 & 44.19 & 72.84 & 54.70 & 59.16 \\

Qwen3.8-flash-next
& 64.82 & 60.46 & 73.26 & 43.90 & 60.61 \\

GLM5.3-flash
& 68.80 & 61.61 & 79.90 & 60.10 & 67.60 \\

\midrule

\multicolumn{6}{l}{\textbf{4B--8B reward models and judges}} \\[-2pt]

Qwen3-4B (official prompts)
& 49.14 & 25.97 & 67.09 & 16.10 & 39.58 \\

Qwen3-8B (official prompts)
& 52.28 & 51.29 & 70.64 & 54.40 & 57.15 \\

Skywork-8B-v0.2
& \textbf{71.48} & 60.81 & 70.00 & 56.50 & 64.70 \\

ArmoRM-8B-v0.1
& 65.41 & 53.39 & 67.64 & 53.60 & 60.01 \\

OpenRubric-4B
& -- & 41.77 & 68.23 & 50.40 & -- \\

OpenRubric-8B
& -- & 40.16 & 68.27 & 52.50 & -- \\

RM-R1-DeepSeek-Distilled-Qwen-7B
& -- & 34.19 & 72.40 & 53.70 & -- \\

RM-R1-Qwen-Instruct-7B
& -- & 62.10 & 70.20 & 54.30 & -- \\

Qwen3-4B-AutoRubric (helpsteer3)
& 53.90 & 18.06 & 54.23 & 50.90 & 44.27 \\

Qwen3-8B-AutoRubric (helpsteer3)
& 57.82 & 52.74 & 62.14 & 50.50 & 55.80 \\

InternLM2-7B-Reward
& 53.39 & 57.74 & 68.02 & 50.10 & 57.31 \\

Llama-3-OffsetBias-RM-8B
& 64.80 & \underline{63.50} & 71.30 & 55.90 & 63.88 \\

JudgeLRM-4B
& 58.00 & 59.35 & 71.21 & 30.90 & 54.87  \\

JudgeLRM-7B
& 40.32 & 31.13 & 50.61 & 31.50 & 38.39 \\

JudgeLRM-8B
& 52.77 & 57.42 & 70.12 & 27.60 & 51.98 \\

\midrule

\multicolumn{6}{l}{\textbf{Multi-Head ProbeRM (ours)}} \\[-2pt]

Qwen3-4B
& 68.19 & 61.45 & \underline{72.96} & \underline{61.50} & \underline{66.02} \\

Qwen3-8B
& \underline{69.51} & \textbf{65.32} & \textbf{73.00} & \textbf{64.40} & \textbf{68.06} \\

\bottomrule
\end{tabular}%
}

\caption{Performance on RewardBench 2, JudgeBench, RM-Bench, and xf-edu-preference. Scores are percentages; higher is better. AVG is the arithmetic mean of the four benchmark scores. Rows labeled ``official prompts'' use each benchmark's official prompt. A dash denotes an unavailable result.}
\label{tab:baseline-performance}
\end{table*}

\subsection{Efficiency Comparison}

\begin{figure*}[!tp]
    \centering
    \includegraphics[width=\textwidth]{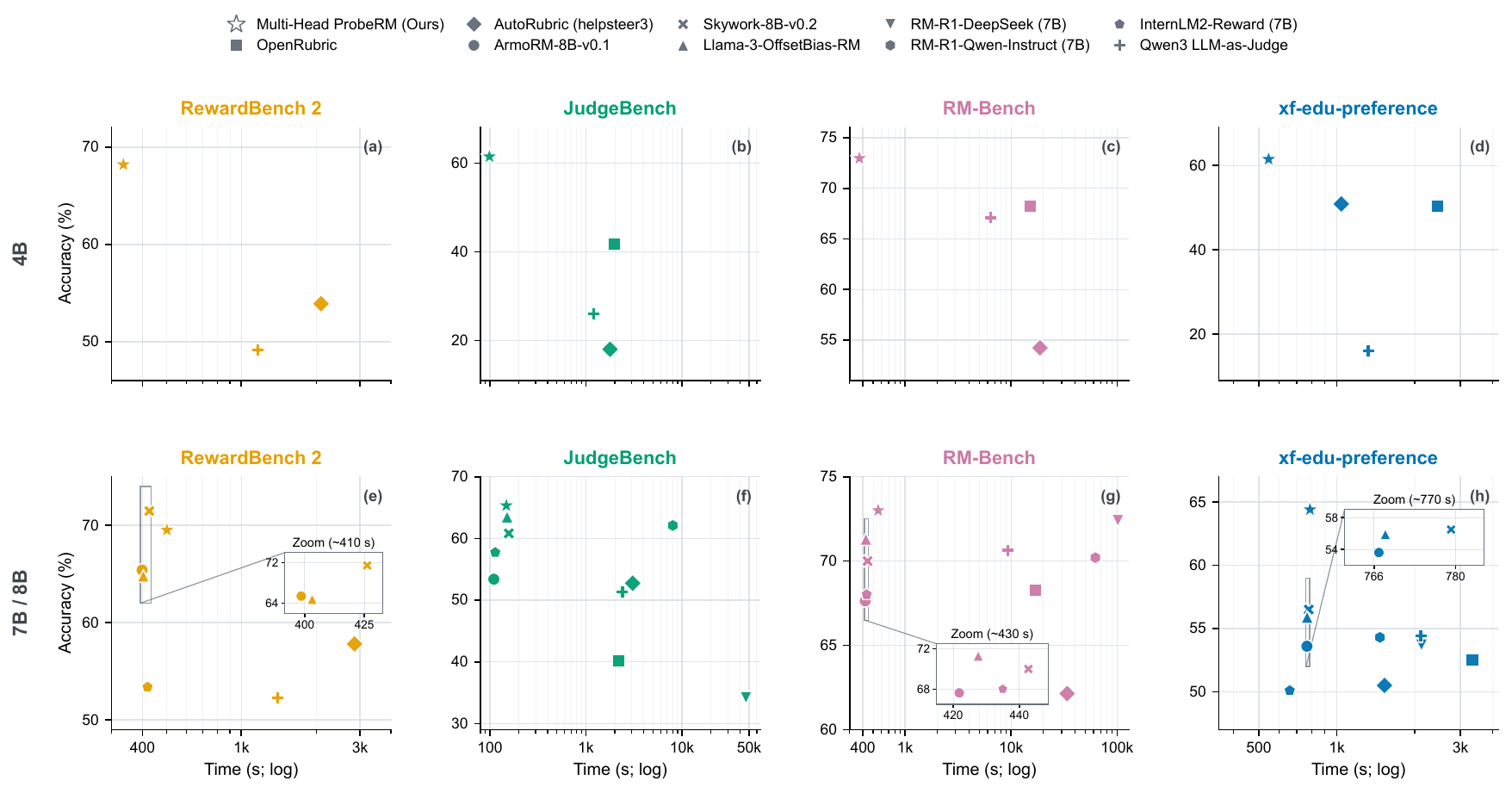}
    \caption{Runtime and benchmark performance. The upper and lower rows compare the 4B and 8B versions of Multi-Head ProbeRM with the indicated baselines. Columns show RewardBench~2, JudgeBench, RM-Bench, and xf-edu-preference. Horizontal axes show total inference time in seconds on a logarithmic scale; vertical axes show benchmark scores in percent. Each model is timed on the full evaluation set of the corresponding benchmark, so the xf-edu-preference panels use the same 1,000 test pairs for accuracy and timing; the exact times are listed in Appendix~\ref{app:inference-time}. Insets magnify the cluster of scalar reward models, which finish RewardBench~2 and RM-Bench in roughly 400--440~s and xf-edu-preference in roughly 660--780~s.} \Description{Eight scatter plots compare runtime and benchmark scores for the 4B and 8B ProbeRM models and their baselines across four benchmarks. The lower panels include magnified views of scalar reward models with runtimes near 410, 430, and 770 seconds.}
    \label{fig:time_accuracy}
\end{figure*}

\label{sec:efficiency}
Figure~\ref{fig:time_accuracy} compares runtime with benchmark accuracy. We time every model on the full evaluation set of each benchmark, namely the 1,865 RewardBench~2 groups, 620 JudgeBench pairs, 11,943 RM-Bench comparisons, and 1,000 xf-edu-preference pairs, under a fixed two-GPU protocol. Appendix~\ref{app:inference-time} reports the per-benchmark times and their totals. Multi-Head ProbeRM computes five dimension scores from one shared backbone encoding of each candidate, without generating rubric or evaluation text.

\paragraph{Comparison with generative judges.}
On xf-edu-preference, ProbeRM takes 545.58~s at 4B and 788.09~s at 8B, whereas OpenRubric takes 2,445.20~s and 3,337.09~s, so ProbeRM is $4.48\times$ and $4.23\times$ faster, respectively. The gap widens on other benchmarks, where rubric and judgment generation dominates the cost: on RM-Bench, OpenRubric needs 15,036.06~s and 16,956.08~s, $40.48\times$ and $30.35\times$ the 371.47~s and 558.68~s of ProbeRM. Against AutoRubric, the per-benchmark speedups at 4B and 8B range from $1.91\times$ and $1.94\times$ on xf-edu-preference to $50.47\times$ and $60.49\times$ on RM-Bench, and the totals are 1,351.23~s and 1,997.19~s vs.\ 23,667.58~s and 41,233.21~s, i.e., $17.52\times$ and $20.65\times$ faster over the full suite. The two reasoning judges are slower still: at 8B, ProbeRM is $311.7\times$ and $54.0\times$ faster than the two RM-R1 variants on JudgeBench, and $181.7\times$ and $111.6\times$ faster than RM-R1-DeepSeek-Distilled and RM-R1-Qwen-Instruct on RM-Bench, respectively, though the margin narrows to $2.69\times$ and $1.86\times$ on xf-edu-preference. Moving rubric supervision into probe heads removes autoregressive judging from reward computation while retaining dimension-level feedback.

\paragraph{Comparison with scalar reward models.}
ProbeRM-4B is the fastest model in the comparison: its total of
1,351.23~s is below every scalar reward model (1,625.86--1,806.10~s), and it also leads on each benchmark individually, taking 335.13~s on RewardBench~2, 99.04~s on JudgeBench, 371.47~s on RM-Bench, and 545.58~s on xf-edu-preference. At 8B, ProbeRM completes the suite in 1,997.19~s,
10.6\% to 22.8\% slower than the scalar models (1,625.86--1,806.10~s) while additionally returning five dimension-level scores, and it remains $4.23\times$ faster than OpenRubric on xf-edu-preference. Scalar models therefore retain a modest timing advantage at the matched 8B scale, but all generative judges in the timing study are slower than ProbeRM at both scales.

\subsection{Directional Dimension Destruction}
\label{sec:directional-destruction}

We test whether each probe head is most sensitive to degradation of its corresponding quality dimension. We randomly sample 200 examples from a mixture of public datasets, excluding overlap with the training set, and use the high-quality response from each example as the original. DeepSeek-4.1-flash edits each original response separately along five dimensions: Accuracy, Compliance, Clarity, Coverage, and Fidelity. Human reviewers check and correct the edits so that each variant primarily degrades the intended dimension. This produces five paired sets of 200 responses, all scored alongside the originals by Qwen3-4B Multi-Head ProbeRM.

For an edit targeting dimension $k$, let $\widetilde{y}_i^{(k)}$ denote the degraded version of response $y_i$ to prompt $q_i$. The mean score decrease on head $j$ is
\begin{equation}
  \Delta_{k,j}=\frac{1}{200}\sum_{i=1}^{200}
  \left[s_j(q_i,y_i)-s_j(q_i,\widetilde{y}_i^{(k)})\right],
  \label{eq:directional-score-drop}
\end{equation}
where a positive value indicates a decrease after degradation. We also measure the target-head decrease rate: the proportion of pairs in which the degraded response receives a lower score on the targeted dimension.

\begin{figure}[!htbp]
  \centering
  \includegraphics[width=\columnwidth]{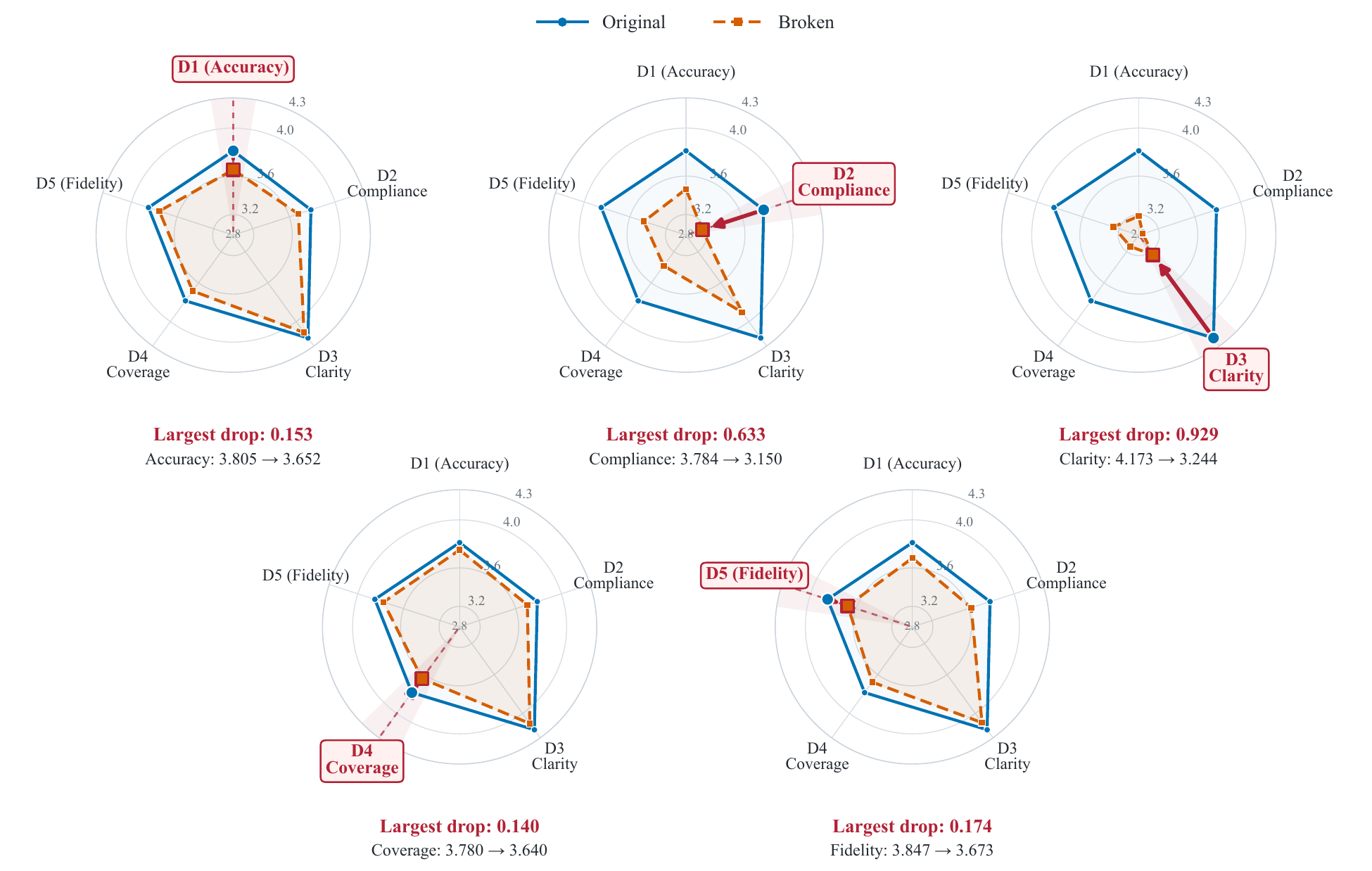}
  \caption{Directional dimension destruction on 200 examples scored by Qwen3-4B Multi-Head ProbeRM. Blue and orange profiles show mean scores before and after targeted degradation. D1--D5 correspond to Accuracy, Compliance, Clarity, Coverage, and Fidelity. Each targeted head has the largest mean score decrease, highlighted in red. Panels share a nonlinear radial scale; annotations report decreases in the original score units.}
  \Description{Five radar charts compare original and degraded responses across five quality dimensions. In each chart, the targeted dimension has the largest mean score decrease.}
  \label{fig:directional-destruction}
\end{figure}

As shown in Figure~\ref{fig:directional-destruction}, the targeted head records the largest mean score decrease in every set. The decreases for Accuracy, Compliance, Clarity, Coverage, and Fidelity are 0.153, 0.633, 0.929, 0.140, and 0.174, respectively. Their decrease rates are 88.0\% (176/200), 95.5\% (191/200), 100.0\% (200/200), 87.5\% (175/200), and 100.0\% (200/200), with no ties. The heads therefore capture the intended degradation both in the mean score profiles and across individual responses. Edits also affect related dimensions: Clarity degradation, for example, reduces Compliance by 0.858 alongside the 0.929 decrease in Clarity. The consistently largest decrease on the targeted head shows that the five outputs distinguish the corresponding aspects of response quality and provide useful dimension-level feedback.
The complete before/after profiles, cross-head decrease matrix, and target-head decrease counts are reported in Appendix~\ref{app:directional-destruction}.

\section{Analysis}
\subsection{Probe-Head Analysis}
\begin{figure*}[!t]
  \centering
  \includegraphics[width=\textwidth]{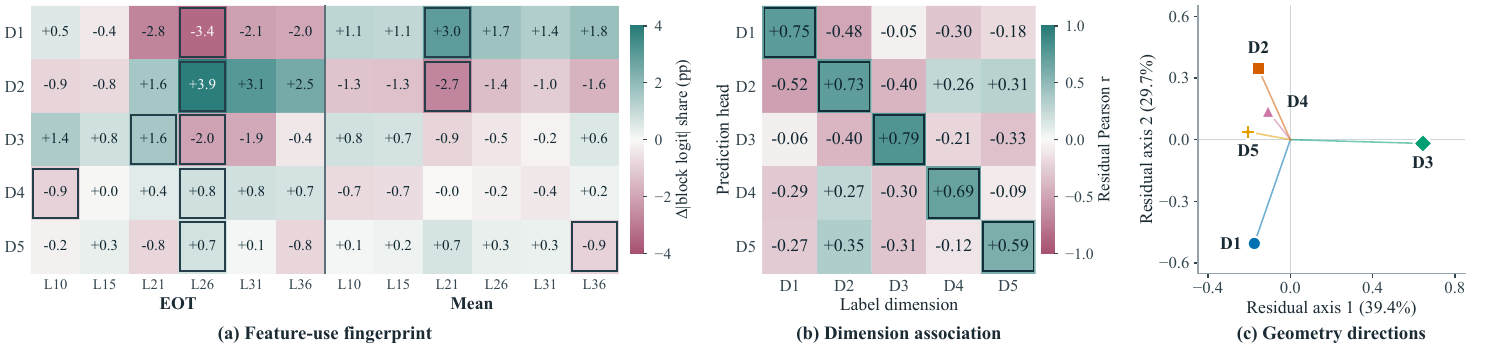}
  \caption{Specialization of the five probe heads. (a) Differences in the normalized share of absolute block-logit contributions between each head and the mean of the other four, measured on 1,200 development responses and expressed in percentage points. The six EOT blocks and six answer-token mean blocks are grouped separately, and boxes mark each head's largest positive and negative differences. (b) Pearson correlations between prediction and teacher-label residuals after regressing each dimension on the mean of its other four dimensions. (c) Two-dimensional projection of normalized probe weights after removing their shared direction.}
  \Description{A three-panel figure shows feature-use differences across EOT and answer-token mean blocks, residual prediction-label correlations, and the projected residual weight directions of five heads.}
  \label{fig:probe-head-analysis}
\end{figure*}

We analyze the five probe heads on 1,200 development-set responses. Figure~\ref{fig:probe-head-analysis}(a) compares each head's normalized share of absolute block-logit contributions against the average of the other four heads. D1 emphasizes layer-21 mean-pooled features (+2.95 percentage points), whereas D2 favors layer-26 EOT features (+3.88 points). D3 shows a relative preference for layer-21 EOT features (+1.57 points), while D4 and D5 exhibit smaller deviations. These patterns reveal different allocations of feature contributions across layers and pooling strategies.

After removing the shared direction from normalized probe weights, Figure~\ref{fig:probe-head-analysis}(c) shows distinct residual directions, particularly for D1 and D3. The two-dimensional projection preserves 69.2\% of the residual weight energy. To examine dimension specificity, we separately regress each prediction and label dimension on the mean of its other four dimensions and correlate the resulting residuals. Figure~\ref{fig:probe-head-analysis}(b) shows that each head retains its strongest residual correlation with its corresponding label, with diagonal Pearson correlations ranging from 0.59 to 0.79. Together, these results support dimension-specific specialization in feature use and scoring alongside shared structure.

\subsection{Ablation Study}
\label{sec:ablation}
We examine how backbone adaptation, feature pooling, layer selection, and dimension-specific probes affect response ranking. Full uses the frozen LoRA-adapted backbone, concatenates EOT and answer-token mean features from six decoder layers, and averages the outputs of five independent probes. Table~\ref{tab:ablation} reports the arithmetic mean of the official scores on RewardBench~2, JudgeBench, RM-Bench, and xf-edu-preference for each configuration at both model scales. Each benchmark receives equal weight; the final AVG column averages the 4B and 8B results.

\begin{table}[t]
    \centering
    \resizebox{\columnwidth}{!}{%
        \begin{tabular}{@{}lccc@{}}
            \toprule
            Configuration & Qwen3-4B & Qwen3-8B & \shortstack{AVG} \\
            \midrule
            Full                & \textbf{66.02\%} &\textbf{ 68.06\%} & \textbf{67.04\%} \\
            w/o LoRA            & 64.68\% & 66.00\% & 65.34\% \\
            LoRA-only           & 61.25\% & 62.46\% & 61.86\% \\
            EOT-only            & 65.31\% & 67.68\% & 66.50\% \\
            Mean-only           & 63.95\% & 65.86\% & 64.91\% \\
            Last-layer only     & 65.18\% & 67.88\% & 66.53\% \\
            Single scalar probe & 65.85\% & 66.98\% & 66.42\% \\
            \bottomrule
        \end{tabular}%
    }
    \caption{Ablation results. Each model column reports the arithmetic mean of its scores on RewardBench~2, JudgeBench, RM-Bench, and xf-edu-preference. AVG averages the two model columns. All scores are percentages; higher is better.}
    \label{tab:ablation}
\end{table}

\paragraph{LoRA adaptation.}
The w/o LoRA variant extracts features from the pretrained backbone without LoRA adaptation. Its four-benchmark mean falls from 66.02\% to 64.68\% at 4B and from 68.06\% to 66.00\% at 8B, corresponding to decreases of 1.34 and 2.06 percentage points. The decline at both scales supports adapting the backbone before freezing its representations for probe training.

The LoRA-only variant is the scalar outcome reward model obtained from the initial Bradley--Terry adaptation stage. Full improves its four-benchmark mean from 61.25\% to 66.02\% at 4B and from 62.46\% to 68.06\% at 8B, gains of 4.77 and 5.60 percentage points. This comparison measures the combined benefit of the additional teacher supervision and rubric readouts on top of the adapted backbone.

\paragraph{Pooling strategy.}
EOT-only retains the end-of-turn features, whereas Mean-only retains the answer-token mean features. Relative to Full, EOT-only lowers the mean by 0.71 points at 4B and 0.38 points at 8B; Mean-only produces larger decreases of 2.07 and 2.20 points. EOT-only outperforms Mean-only at both scales, indicating a stronger standalone ranking signal in the EOT representation. Combining the two feature types yields the highest mean scores, suggesting that answer-token mean features complement the EOT features.

\paragraph{Layer selection.}
Last-layer only restricts feature extraction to the final decoder layer instead of combining six layers. It yields 65.18\% at 4B and 67.88\% at 8B, falling below Full by 0.84 and 0.18 points. The multi-layer representation improves the four-benchmark mean at both scales, with a larger benefit at 4B. This result supports retaining intermediate-layer features alongside the final-layer representation.

\paragraph{Dimension-specific probes.}
Single scalar probe uses the same LoRA checkpoint and the same multi-layer EOT and answer-token mean features as Full, but predicts one scalar target: the mean of the five teacher-derived dimension targets, $\bar{t}_i=\frac{1}{5}\sum_{k=1}^{5}t_{ik}$. Full fits each dimension separately before averaging its predictions. The scalar probe yields 65.85\% at 4B and 66.98\% at 8B, decreases of 0.17 and 1.08 points relative to Full. The advantage of the five-head configuration is larger at 8B, while the 4B results remain close. Across the two scales, Full reaches 67.04\% compared with 66.42\% for the single scalar probe. Thus, the five-head design retains interpretable dimension-level scores and achieves higher average ranking performance in this comparison.

\section{Conclusion}

We presented Multi-Head ProbeRM, which connects a shared rubric derived from preference data to five interpretable quality scores and a scalar ranking reward. The rubric defines the offline supervision, while all five scores are computed from one frozen prompt--response representation. After LoRA adaptation of Qwen3 with a Bradley--Terry loss on source preference pairs, five independent probes learn Accuracy, Compliance, Clarity, Coverage, and Fidelity from the mean of two teachers' integer scores. The arithmetic mean of the five probe outputs provides a scalar reward with an explicit contribution from each dimension. This design retains rubric-aligned feedback and computes all five scores from the prompt and response alone.

Experiments on RewardBench~2, JudgeBench, RM-Bench, and xf-edu-preference demonstrate strong ranking performance, with four-benchmark averages of 66.02\% and 68.06\% for the 4B and 8B models. The 8B model reaches 64.40\% on xf-edu-preference, exceeding the scalar and generative baselines reported for this test set. In the matched timing study, OpenRubric takes 4.48$\times$ and 4.23$\times$ as long as our 4B and 8B models on xf-edu-preference, where all 1,000 pairs receive human preference labels, and AutoRubric takes 17.52$\times$ and 20.65$\times$ as long over the full four-benchmark suite. Targeted degradation and probe-head analyses further support the dimension specificity of the learned scores. These findings show that offline rubric supervision can combine interpretable feedback, efficient inference, and accurate preference ranking in a single reward model.

\section{Limitations}
\label{sec:limitations}

The shared rubric is induced from 70 preference pairs in the General domain of HelpSteer3-Preference, so the five dimensions reflect that narrow sample; code and mathematics are covered only through the generalization into 25 criteria, and the criteria are fixed rather than adapted per query. The teacher model GPT-6.1-Sol is a closed-source hosted API, so its scores cannot be audited or reproduced as model versions change. Averaging two integer scores places targets on a 0.5-point grid with at most nine values per dimension, which limits supervision resolution. AVG averages four benchmarks of differing size and difficulty, from 620 JudgeBench pairs to 11,943 RM-Bench comparisons, and is a descriptive summary rather than a calibrated measure of general judging ability. 

\section*{Ethics and Privacy Statement}
The xf-edu-preference test set is derived from real educational applications and underwent privacy protection and de-identification before use. Professional teachers annotate preferences under established rules, with model identities hidden and candidate order randomized. All 1,000 pairs are reserved for final evaluation and are excluded from rubric construction, LoRA adaptation, probe training. Appendix~\ref{app:edu-preference} describes the data and annotation protocol.

\bibliographystyle{ACM-Reference-Format}
\bibliography{Reference}

\clearpage
\appendix

\section{Rubric Definitions}
\label{app:general-rubrics}

This appendix lists the 25 criteria used to construct the offline teacher
supervision.  The criteria are generalized from 29 core rubrics derived from
the General domain of HelpSteer3-Preference and are organized into five
dimensions, with five criteria per dimension.  The persistent dimension names
and identifiers below are also used in the training-data exports.

The short names used in the abstract, main text, figures, and tables map to
the expanded rubric names as follows:
\begin{description}
  \item[\textbf{D1: Accuracy}] Factual Accuracy \& Context Consistency.
  \item[\textbf{D2: Compliance}] Strict Compliance with Requirements.
  \item[\textbf{D3: Clarity}] Clarity \& Structured Presentation.
  \item[\textbf{D4: Coverage}] Comprehensive \& Coherent Coverage.
  \item[\textbf{D5: Fidelity}] Narrative \& Context Fidelity.
\end{description}
Both naming forms refer to the same five dimensions and the same criteria.

\subsection{Factual Accuracy \& Context Consistency (D1)}
\label{app:rubric-accuracy}

\begin{description}
  \item[\textbf{D1-C1.}] Ranks higher the response that states the required facts, concepts, procedures, and relationships accurately, with no material factual errors.
  \item[\textbf{D1-C2.}] Ranks higher the response that distinguishes established information from interpretation, uncertainty, developing law, speculation, or hypothetical claims when that distinction matters.
  \item[\textbf{D1-C3.}] Ranks higher the response whose numbers, dates, calculations, timings, and logistical details are internally consistent and compatible with the stated circumstances.
  \item[\textbf{D1-C4.}] Ranks lower the response that invents unsupported sources, capabilities, quotations, historical claims, technical features, or precise figures, especially when presented as certain.
  \item[\textbf{D1-C5.}] Ranks higher the response that uses domain-specific terminology correctly and does not confuse closely related entities, categories, roles, or mechanisms.
\end{description}

\subsection{Strict Compliance with Requirements (D2)}
\label{app:rubric-adherence}

\begin{description}
  \item[\textbf{D2-C1.}] Ranks higher the response that directly fulfills every substantive part of the request rather than answering a nearby question or substituting general commentary.
  \item[\textbf{D2-C2.}] Ranks higher the response that obeys explicit constraints on quantity, length, structure, wording, formatting, parameters, and output type.
  \item[\textbf{D2-C3.}] Ranks higher the response that follows requested scope, audience, tone, perspective, and level of detail without adding irrelevant material or omitting required elements.
  \item[\textbf{D2-C4.}] Ranks higher the response that provides the requested practical deliverable, such as steps, examples, options, calculations, prompts, or a usable draft, when one is requested.
  \item[\textbf{D2-C5.}] Ranks lower the response that refuses a benign request without justification, adds disallowed content, violates specified boundaries, or treats optional material as a replacement for core requirements.
\end{description}

\subsection{Clarity \& Structured Presentation (D3)}
\label{app:rubric-clarity}

\begin{description}
  \item[\textbf{D3-C1.}] Ranks higher the response that presents information in a logical order with clear grouping, headings, transitions, or other structure appropriate to the task.
  \item[\textbf{D3-C2.}] Ranks higher the response that uses precise, accessible wording and makes the main answer or conclusion easy to identify.
  \item[\textbf{D3-C3.}] Ranks higher the response that explains specialized terms, assumptions, and reasoning sufficiently for the intended audience without relying on vague assertions.
  \item[\textbf{D3-C4.}] Ranks higher the response that is concise relative to the request, removing repetition, filler, tangents, and unnecessary disclaimers while retaining useful distinctions.
  \item[\textbf{D3-C5.}] Ranks higher the response whose formatting, paragraphs, lists, dialogue, or prose are easy to scan and do not create ambiguity about which information belongs together.
\end{description}

\subsection{Comprehensive \& Coherent Coverage (D4)}
\label{app:rubric-completeness}

\begin{description}
  \item[\textbf{D4-C1.}] Ranks higher the response that covers the major requested subtopics and relevant decision factors, with no material omission of information needed to use the answer.
  \item[\textbf{D4-C2.}] Ranks higher the response that supplies concrete, sufficiently developed details rather than merely naming categories, options, locations, arguments, or procedures.
  \item[\textbf{D4-C3.}] Ranks higher the response that connects its parts into a consistent whole, so recommendations, explanations, examples, and conclusions support rather than contradict one another.
  \item[\textbf{D4-C4.}] Ranks higher the response that gives appropriate breadth while prioritizing the core task, distinguishing essential elements from optional additions and avoiding impractical overloading.
  \item[\textbf{D4-C5.}] Ranks higher the response that offers actionable reasoning or development guidance where appropriate, enabling the user to narrow, apply, compare, or extend the information.
\end{description}

\subsection{Narrative \& Context Fidelity (D5)}
\label{app:rubric-fidelity}

\begin{description}
  \item[\textbf{D5-C1.}] Ranks higher the response that preserves the supplied setting, premises, roles, relationships, chronology, and other context-specific constraints without contradiction.
  \item[\textbf{D5-C2.}] Ranks higher the response that maintains the requested voice, tone, genre, point of view, and interpersonal dynamics throughout.
  \item[\textbf{D5-C3.}] Ranks higher a creative continuation or transformation that develops the particular setup, mystery, conflict, or emotional premise instead of replacing it with a generic scenario.
  \item[\textbf{D5-C4.}] Ranks higher the response that keeps participants, stakeholders, or fictional characters acting consistently with their established perspectives and circumstances while allowing meaningful development.
  \item[\textbf{D5-C5.}] Ranks higher the response that respects whether the task is factual, fictional, hypothetical, or role-based and does not break immersion or introduce irrelevant meta-commentary.
\end{description}

\section{Theoretical Analysis of EOT and Answer-Only Mean Concatenation}
\label{app:pooling-theory}

We analyze why concatenating the end-of-turn (EOT) representation with
the answer-only mean can retain complementary information for reward
prediction. The analysis separates an architectural motivation from
information-preservation guarantees and conditions for strict predictive
improvement.

\subsection{Representation and Architectural Motivation}
\label{app:pooling-setup}

Let $X$ denote the model input consisting of a prompt, an answer, and the
assistant EOT token. All representations refer to the retained input
after the truncation policy described in Section~\ref{sec:reward-model}.
Let $\mathcal{A}$ be the set of retained answer-token positions, with
$n=|\mathcal{A}|>0$, and let $h_i^{(\ell)}\in\mathbb{R}^{H}$ be the hidden
state at position $i$ in layer $\ell$ of the frozen causal Transformer.
For the selected layers $\mathcal{L}=\{10,15,21,26,31,36\}$, define
\begin{equation}
\begin{aligned}
E &= \operatorname{concat}_{\ell\in\mathcal{L}}
     h_{\mathrm{EOT}}^{(\ell)}, \\
M &= \operatorname{concat}_{\ell\in\mathcal{L}}
     \left(\frac{1}{n}\sum_{i\in\mathcal{A}}h_i^{(\ell)}\right),
\qquad Z=[E;M].
\end{aligned}
\label{eq:pooling-representations}
\end{equation}
When truncation removes the EOT marker, $h_{\mathrm{EOT}}^{(\ell)}$ in
this definition denotes the last retained valid readout used by the
implementation. The feature vector in Eq.~\ref{eq:probe-feature} is an
interleaved ordering of the same coordinates; it differs from $Z$ only by
a fixed permutation. Thus, $E,M\in\mathbb{R}^{p}$, where
$p=H|\mathcal{L}|$, and $Z\in\mathbb{R}^{2p}$.
Let $Y$ denote the target reward or vector of rubric scores.

The causal attention mask allows the EOT position to access the entire
retained prompt--answer prefix. This makes $E$ a sequence-terminal readout
with a global receptive field. In contrast, $M$ directly aggregates
contextualized features from all answer positions, exposing evidence
distributed across the answer to the prediction head. These properties
motivate combining a terminal readout with a distributed token readout.
The global receptive field does not itself imply that $E$ is a sufficient
statistic for $Y$. Likewise, mean pooling is a lossy aggregation of
contextualized token states, rather than a lossless representation of
local details. The results below specify what concatenation guarantees
and when complementary evidence yields a strict gain.

\subsection{Preservation of Reward-Relevant Information}
\label{app:pooling-information}

\begin{proposition}[Information preservation and conditional gain]
\label{prop:pooling-information}
Suppose the mutual information $I(Y;E,M)$ is finite, as is the case for
finite-valued rubric labels~\cite{cover2006elements}. Then
\begin{equation}
I(Y;Z)\geq\max\{I(Y;E),I(Y;M)\}.
\label{eq:pooling-information-bound}
\end{equation}
The additional reward-relevant information relative to EOT alone is
\begin{equation}
I(Y;Z)-I(Y;E)=I(Y;M\mid E).
\label{eq:pooling-information-gain}
\end{equation}
Consequently, the inequality relative to EOT is strict if and only if
$Y$ and $M$ are not conditionally independent given $E$.
\end{proposition}

\begin{proof}
Since $Z=[E;M]$, the mutual information chain rule gives
\begin{equation}
\begin{aligned}
I(Y;Z)
&=I(Y;E)+I(Y;M\mid E) \\
&=I(Y;M)+I(Y;E\mid M).
\end{aligned}
\end{equation}
Both conditional mutual information terms are nonnegative, proving
Eq.~\ref{eq:pooling-information-bound} and
Eq.~\ref{eq:pooling-information-gain}. Moreover,
$I(Y;M\mid E)=0$ if and only if $Y$ and $M$ are conditionally
independent given $E$, which establishes the strictness condition.
\end{proof}

The quantity $I(Y;M\mid E)$ measures the scoring information supplied by
the answer-token aggregate after the EOT representation is already
known. Complementarity therefore does not require the two readouts to
be independent or to encode disjoint information. This information-based
reading follows the standard treatment of probing as the estimation of
the mutual information between a representation and a target
attribute~\cite{voita2020information}.

\subsection{Optimal Risk Within the Linear Probe Class}
\label{app:pooling-probe-risk}

The information guarantee has a corresponding consequence for the
single-output linear heads used in this paper.

\begin{proposition}[Probe-class risk does not increase]
\label{prop:pooling-probe-risk}
Let $\mathcal{H}_E$, $\mathcal{H}_M$, and $\mathcal{H}_Z$ be probe
classes taking $E$, $M$, and $Z$ as input, respectively. Suppose they
share the same affine output architecture, including the same output
transformation, except for the input dimension, and that the first
operation is an unrestricted affine map. For any nonnegative measurable
loss $\ell$, define
\begin{equation}
R_{\mathcal{H}_U}^{\star}
=\inf_{f\in\mathcal{H}_U}\mathbb{E}[\ell(Y,f(U))],
\qquad U\in\{E,M,Z\}.
\end{equation}
Then
\begin{equation}
R_{\mathcal{H}_Z}^{\star}
\leq\min\left\{
R_{\mathcal{H}_E}^{\star},R_{\mathcal{H}_M}^{\star}\right\}.
\label{eq:pooling-probe-risk}
\end{equation}
\end{proposition}

\begin{proof}
Consider any EOT-only probe whose first affine map is $W_E E+b$.
Construct a concatenated-input probe with first-layer weights
$[W_E\;0]$ and the same bias and output transformation. Its affine
preactivation is
\begin{equation}
[W_E\;0]\begin{bmatrix}E\\M\end{bmatrix}+b=W_E E+b,
\end{equation}
so it reproduces the EOT-only probe on every input. Setting the EOT
block to zero gives the corresponding inclusion for $\mathcal{H}_M$.
Taking infima of the expected loss proves
Eq.~\ref{eq:pooling-probe-risk}.
\end{proof}

This argument applies to the sigmoid-bounded affine heads used in the
paper and therefore applies to their Smooth~L1 training objective. The
train-only coordinatewise standardization is fixed during probe fitting;
the zero-weight construction remains valid after this deterministic
preprocessing. The result concerns optimal population risk within the
specified probe class. A finite-sample trained head can show a smaller,
equal, or indistinguishable empirical improvement because estimation and
optimization are separate effects. Control tasks measure how much of a
probe's performance reflects the representation rather than the probe's
own capacity~\cite{hewitt2019designing}, which is the reason the
comparisons in Table~\ref{tab:ablation} vary one design factor at a time.

\subsection{Conditions for Strict Improvement Under Squared Loss}
\label{app:pooling-strict-gain}

To characterize when the additional readout improves prediction
strictly, we examine the Bayes risk under squared loss.

\begin{proposition}[Exact Bayes-risk reduction]
\label{prop:pooling-bayes-risk}
Let $Y\in\mathbb{R}^{k}$ satisfy
$\mathbb{E}\|Y\|_2^2<\infty$. Define
\begin{equation}
\begin{aligned}
\mu_E&=\mathbb{E}[Y\mid E],
&\mu_Z&=\mathbb{E}[Y\mid E,M],\\
\mathcal{R}^{\star}(U)
&=\inf_{f}\mathbb{E}\|Y-f(U)\|_2^2,
& U&\in\{E,Z\},
\end{aligned}
\end{equation}
where the infimum is over measurable predictors. Then
\begin{equation}
\mathcal{R}^{\star}(E)-\mathcal{R}^{\star}(Z)
=\mathbb{E}\|\mu_Z-\mu_E\|_2^2\geq0.
\label{eq:pooling-bayes-risk}
\end{equation}
The reduction is strict if and only if
$\mathbb{P}(\mu_Z\neq\mu_E)>0$.
\end{proposition}

\begin{proof}
The conditional expectation minimizes expected squared loss, so
$\mathcal{R}^{\star}(E)=\mathbb{E}\|Y-\mu_E\|_2^2$ and
$\mathcal{R}^{\star}(Z)=\mathbb{E}\|Y-\mu_Z\|_2^2$.
Since both $\mu_E$ and $\mu_Z$ are measurable with respect to $(E,M)$,
\begin{equation}
\begin{aligned}
&\mathbb{E}\langle Y-\mu_Z,\mu_Z-\mu_E\rangle\\
&\quad=\mathbb{E}\left[
\left\langle\mathbb{E}[Y-\mu_Z\mid E,M],
\mu_Z-\mu_E\right\rangle\right]=0.
\end{aligned}
\end{equation}
Expanding $Y-\mu_E=(Y-\mu_Z)+(\mu_Z-\mu_E)$ therefore gives
\begin{equation}
\mathbb{E}\|Y-\mu_E\|_2^2
=\mathbb{E}\|Y-\mu_Z\|_2^2
+\mathbb{E}\|\mu_Z-\mu_E\|_2^2.
\end{equation}
Rearranging yields Eq.~\ref{eq:pooling-bayes-risk}. The expected squared
norm is positive exactly when its argument is nonzero with positive
probability.
\end{proof}

An additive conditional-mean model makes the role of residual
answer-token evidence explicit.

\begin{corollary}[Gain from residual answer-token evidence]
\label{cor:pooling-residual-gain}
Suppose $E$, $M$, and $Y$ have finite second moments and
\begin{equation}
Y=AE+BM+\varepsilon,
\qquad \mathbb{E}[\varepsilon\mid E,M]=0,
\label{eq:pooling-additive-model}
\end{equation}
where $A,B\in\mathbb{R}^{k\times p}$. Let
$m(E)=\mathbb{E}[M\mid E]$ and
$\Sigma_{M\mid E}=\operatorname{Cov}(M\mid E)$. Then
\begin{equation}
\begin{aligned}
\mathcal{R}^{\star}(E)-\mathcal{R}^{\star}(Z)
&=\mathbb{E}\|B(M-m(E))\|_2^2\\
&=\operatorname{tr}\!\left(
B\,\mathbb{E}[\Sigma_{M\mid E}]\,B^\top
\right).
\end{aligned}
\label{eq:pooling-residual-gain}
\end{equation}
In particular, concatenation strictly reduces Bayes squared error if
$\mathbb{P}(B(M-m(E))\neq0)>0$.
\end{corollary}

\begin{proof}
Taking conditional expectations in
Eq.~\ref{eq:pooling-additive-model} gives
$\mu_Z=AE+BM$ and $\mu_E=AE+Bm(E)$, where the second equality follows
from the tower property. Substitution into
Eq.~\ref{eq:pooling-bayes-risk} gives the first equality in
Eq.~\ref{eq:pooling-residual-gain}. For
$V=M-m(E)$, the conditional covariance identity yields
$\mathbb{E}[VV^\top]=\mathbb{E}[\Sigma_{M\mid E}]$.
Using $\|BV\|_2^2=\operatorname{tr}(BVV^\top B^\top)$ proves the
second equality and the stated strictness condition.
\end{proof}

The residual $M-m(E)$ is the component of the answer-token aggregate
that cannot be predicted from EOT in the conditional-mean sense. The
matrix $B$ selects directions relevant to the target scores. Hence,
under Eq.~\ref{eq:pooling-additive-model}, a strict improvement occurs
when this residual contains variation along scoring-relevant directions.
Positive conditional mutual information can also reflect changes in the
conditional distribution that leave its mean unchanged; strict
information gain and strict squared-error improvement therefore have
distinct conditions.

Together, these results support concatenating EOT and answer-only mean
as a way to preserve both readouts and exploit complementary scoring
evidence when present. The squared-loss identities characterize Bayes
predictors, whereas Proposition~\ref{prop:pooling-probe-risk} applies to
the specified probe architecture and its actual loss. The gains obtained
by a trained probe additionally depend on finite-sample estimation and
optimization.

\section{Full Directional Dimension-Destruction Results}
\label{app:directional-destruction}

This appendix reports the complete mean-score profiles and cross-head score
decreases for the directional dimension-destruction experiment in
Section~\ref{sec:directional-destruction}. We use the same 200 paired
examples for each targeted degradation and the same Qwen3-4B
Multi-Head ProbeRM checkpoint. A positive entry in the decrease table
means that the score is lower after the edit. The five degradation sets
are constructed independently, so each row compares the same 200 original
responses with one targeted variant per response.

\begin{table*}[!htbp]
\centering
\caption{Mean probe scores before and after each targeted degradation. Each
row contains the mean score over 200 original responses or their degraded
variants. D1--D5 denote Accuracy, Compliance, Clarity, Coverage, and
Fidelity, respectively.}
\label{tab:directional-before-after}
\small
\setlength{\tabcolsep}{5pt}
\renewcommand{\arraystretch}{1.1}
\begin{tabular*}{\textwidth}{@{\extracolsep{\fill}}llrrrrr@{}}
\toprule
Targeted degradation & State & D1 & D2 & D3 & D4 & D5 \\
\midrule
D1 (Accuracy) & Before & 3.804769 & 3.783595 & 4.173167 & 3.780306 & 3.846633 \\
              & After  & 3.651558 & 3.676392 & 4.115316 & 3.681942 & 3.751638 \\
D2 (Compliance) & Before & 3.804769 & 3.783595 & 4.173167 & 3.780306 & 3.846633 \\
                 & After  & 3.473048 & 3.150470 & 3.899404 & 3.394231 & 3.457312 \\
D3 (Clarity) & Before & 3.804769 & 3.783595 & 4.173167 & 3.780306 & 3.846633 \\
              & After  & 3.175258 & 2.925770 & 3.243761 & 3.107191 & 3.268804 \\
D4 (Coverage) & Before & 3.804769 & 3.783595 & 4.173167 & 3.780306 & 3.846633 \\
               & After  & 3.745464 & 3.698780 & 4.106718 & 3.640244 & 3.770653 \\
D5 (Fidelity) & Before & 3.804769 & 3.783595 & 4.173167 & 3.780306 & 3.846633 \\
               & After  & 3.679525 & 3.624143 & 4.097332 & 3.675415 & 3.672925 \\
\bottomrule
\end{tabular*}
\end{table*}

\begin{table}[!htbp]
\centering
\caption{Mean score decrease $\Delta_{k,j}$ for each targeted degradation
$k$ and probe head $j$. The diagonal entry in every row is the largest
decrease in that row.}
\label{tab:directional-decrease-matrix}
\small
\setlength{\tabcolsep}{4pt}
\renewcommand{\arraystretch}{1.1}
\begin{tabular}{lrrrrr}
\toprule
Target & D1 & D2 & D3 & D4 & D5 \\
\midrule
D1 (Accuracy) & 0.153211 & 0.107203 & 0.057850 & 0.098364 & 0.094995 \\
D2 (Compliance) & 0.331722 & 0.633125 & 0.273762 & 0.386075 & 0.389322 \\
D3 (Clarity) & 0.629512 & 0.857825 & 0.929406 & 0.673115 & 0.577829 \\
D4 (Coverage) & 0.059305 & 0.084816 & 0.066449 & 0.140062 & 0.075981 \\
D5 (Fidelity) & 0.125245 & 0.159453 & 0.075835 & 0.104891 & 0.173708 \\
\bottomrule
\end{tabular}
\end{table}

\begin{table}[!htbp]
\centering
\caption{Target-head decrease rates over the 200 paired examples. No pair
is tied on the targeted head.}
\label{tab:directional-decrease-rate}
\small
\setlength{\tabcolsep}{5pt}
\renewcommand{\arraystretch}{1.1}
\begin{tabular}{lrr}
\toprule
Targeted dimension & Count & Rate \\
\midrule
D1 (Accuracy) & 176/200 & 88.0\% \\
D2 (Compliance) & 191/200 & 95.5\% \\
D3 (Clarity) & 200/200 & 100.0\% \\
D4 (Coverage) & 175/200 & 87.5\% \\
D5 (Fidelity) & 200/200 & 100.0\% \\
\bottomrule
\end{tabular}
\end{table}

The diagonal entries of Table~\ref{tab:directional-decrease-matrix} are
the largest values in their respective rows, matching the qualitative
profile in Figure~\ref{fig:directional-destruction}. The off-diagonal
entries quantify spillover between dimensions. For example, a Clarity
degradation decreases the Compliance head by 0.857825 while decreasing
the Clarity head by 0.929406. Thus, the experiment measures targeted
sensitivity together with the related-dimension effects induced by each
edit.

\section{Educational Preference Dataset}
\label{app:edu-preference}

\paragraph{Source and composition.}
The xf-edu-preference test set contains 250 pairs from each of four application scenarios: Chinese language arts AI analysis, Chinese language arts interactive video, English AI analysis, and English interactive video. The subject labels refer to Chinese language arts and English education. Questions and candidate responses originate from real educational applications, where the responses are produced by educational language models. Each evaluation item presents a dialogue context and two candidate responses in text form. The interactive-video scenarios are evaluated through their dialogue text. The data underwent privacy protection and de-identification before use.

\paragraph{Teacher preference annotation.}
Professional teachers with expertise in the corresponding subjects follow established annotation rules to compare the two responses in their dialogue context and select the overall preferred response. The originating model identities are hidden, and the candidate presentation order is randomized. The annotation task requires a choice between the two candidates and does not offer a tie label. Each pair therefore has a single preferred response as its reference label for evaluation.

\paragraph{Accuracy and failure handling.}
Let $c_i=1$ if the evaluator returns a valid strict preference matching the teacher label for pair $i$, and $c_i=0$ otherwise. Accuracy is
\begin{equation}
  \mathrm{Acc}_{\mathrm{edu}} =
  \frac{100}{1{,}000}\sum_{i=1}^{1{,}000}c_i\quad(\%).
  \label{eq:edu-accuracy}
\end{equation}
For scalar reward models, a correct selection requires the teacher-preferred response to receive a strictly higher score than the other candidate. Equal scores count as errors. Refusals and judgments that cannot be parsed into a valid preference also count as errors and remain in the denominator. Accuracy and inference time are measured on the same 1,000 pairs.

\paragraph{Test isolation.}
All 1,000 pairs are used exclusively for final testing. They are excluded from rubric construction, LoRA adaptation, probe training, and hyperparameter tuning or checkpoint selection. Their teacher preference labels are used to evaluate the completed models.

\section{Inference-Time Results}
\label{app:inference-time}
Table~\ref{tab:inference-time} lists the inference times underlying Section~\ref{sec:efficiency}. Each entry is the wall-clock time for scoring the complete evaluation set of the corresponding benchmark on the fixed two-GPU protocol described in Section~\ref{sec:experiment-settings}. Because the benchmarks differ in size (from 620 JudgeBench pairs to 11,943 RM-Bench comparisons), times are comparable within a column but not across columns; the Total column sums the four columns for models measured on all of them. The xf-edu-preference column uses exactly the same 1,000 pairs as the accuracy evaluation, with 250 pairs from each of the four scenarios.

\begin{table*}[!htbp]
\centering
\caption{Wall-clock inference time in seconds (lower is better) on the full evaluation set of each benchmark: 1,865 RewardBench~2 groups, 620 JudgeBench pairs, 11,943 RM-Bench comparisons, and 1,000 xf-edu-preference pairs. Total is the sum of the four benchmark times. A dash denotes an unavailable value.}
\label{tab:inference-time}
\small
\setlength{\tabcolsep}{4pt}
\renewcommand{\arraystretch}{1.12}
\begin{tabular*}{\textwidth}{@{\extracolsep{\fill}}lrrrrr@{}}
\toprule
Model
& \shortstack[r]{RewardBench 2\\(s)}
& \shortstack[r]{JudgeBench\\(s)}
& \shortstack[r]{RM-Bench\\(s)}
& \shortstack[r]{xf-edu-preference\\(s)}
& \shortstack[r]{Total\\(s)} \\
\midrule

\multicolumn{6}{l}{\textbf{4B--8B reward models and judges}} \\[-2pt]

Qwen3-4B (official prompts)
& 1,165.635 & 1,207.707 & 6,442.167 & 1,322.746 & 10,138.255 \\

Qwen3-8B (official prompts)
& 1,400.628 & 2,402.638 & 9,364.274 & 2,109.805 & 15,277.345 \\

Skywork-8B-v0.2
& 426.263 & 157.797 & 442.809 & 779.228 & 1,806.097 \\

ArmoRM-8B-v0.1
& 398.623 & 110.122 & 421.776 & 766.835 & 1,697.356 \\

OpenRubric-4B
& -- & 1,989.691 & 15,036.059 & 2,445.196 & -- \\

OpenRubric-8B
& -- & 2,203.989 & 16,956.080 & 3,337.090 & -- \\

RM-R1-DeepSeek-Distilled-Qwen-7B
& -- & 46,307.544 & 101,521.220 & 2,122.353 & -- \\

RM-R1-Qwen-Instruct-7B
& -- & 8,020.407 & 62,340.439 & 1,466.373 & -- \\

Qwen3-4B-AutoRubric (helpsteer3)
& 2,093.799 & 1,785.725 & 18,747.525 & 1,040.529 & 23,667.578 \\

Qwen3-8B-AutoRubric (helpsteer3)
& 2,852.947 & 3,060.682 & 33,791.755 & 1,527.823 & 41,233.207 \\

InternLM2-7B-Reward
& 419.133 & 113.793 & 434.900 & 658.038 & 1,625.864 \\

Llama-3-OffsetBias-RM-8B
& 403.081 & 151.185 & 427.449 & 767.933 & 1,749.648 \\

\midrule

\multicolumn{6}{l}{\textbf{Multi-Head ProbeRM (ours)}} \\[-2pt]

Qwen3-4B
& 335.133 & 99.044 & 371.472 & 545.578 & 1,351.227 \\

Qwen3-8B
& 501.842 & 148.573 & 558.676 & 788.094 & 1,997.185 \\

\bottomrule
\end{tabular*}
\end{table*}

\end{document}